\pdfoutput=1
\documentclass[lettersize,journal]{IEEEtran}
\usepackage{amsmath,amsfonts}
\usepackage{algorithmic}
\usepackage{array}
\usepackage[caption=false,font=normalsize,labelfont=sf,textfont=sf]{subfig}
\usepackage{textcomp}
\usepackage{stfloats}
\usepackage{url}
\usepackage{verbatim}
\usepackage{graphicx}
\usepackage{multirow}
\usepackage{multicol}
\usepackage{booktabs}
\usepackage{hyperref}
\usepackage{xcolor}

\newlength{\CopyrightNoticeYShift}
\def\BibTeX{{\rm B\kern-.05em{\sc i\kern-.025em b}\kern-.08em
    T\kern-.1667em\lower.7ex\hbox{E}\kern-.125emX}}
\usepackage{balance}

\definecolor{softred}{hsb}{0,0,0}
\definecolor{softblue}{hsb}{0.60,0.40,0.90}

\newcommand{\IEEECopyrightNotice}{
\textcolor{black!65}{
\footnotesize
\textcopyright~2026 IEEE. Personal use of this material is permitted. Permission from IEEE must be obtained for all other uses, in any current or future media, including reprinting/republishing this material for advertising or promotional purposes, creating new collective works, for resale or redistribution to servers or lists, or reuse of any copyrighted component of this work in other works. This article has been accepted for publication in
\emph{IEEE Transactions on Image Processing}, vol.~35, pp.~7670--7684, 2026. DOI: 10.1109/TIP.2026.3710486
}
}

\makeatletter
\def\ps@IEEEtitlepagestyle{
  \def\@oddhead{
  \raisebox{\CopyrightNoticeYShift}[0pt][0pt]{
    \parbox[t]{0.98\textwidth}{\footnotesize \IEEECopyrightNotice}
    }
    \hfil\thepage}
  \def\@evenhead{
    \thepage\hfil
    \parbox[t]{0.98\textwidth}{\footnotesize \IEEECopyrightNotice}}
  \let\@oddfoot\@empty
  \let\@evenfoot\@empty
}
\makeatother

\begin{document}

\title{\textcolor{softred}{CGGS: Consistency-Augmented Geometric Gaussian Splatting for Ego-Centric 3D Scene Generation}}

\author{Zhenyu Sun,~Xiaohan Zhang,~Qi Liu$^{\dagger}$,~\IEEEmembership{Senior~Member,~IEEE}, and Huan Wang$^{\dagger}$,~\IEEEmembership{Senior~Member,~IEEE}

\thanks{
This work was supported in part by the Young Scientists Fund of the National Natural Science Foundation of China (NSFC) under Grant 62506305, in part by Zhejiang Leading Innovative and Entrepreneur Team Introduction Program under Grant 2024R01007, in part by the Key Research and Development Program of Zhejiang Province under Grant 2025C01026, in part by the Scientific Research Project of Westlake University under Grant WU2025WF003, in part by the GuangJuYingCai (GJYC) Program of Guangzhou under Grant 2024D01J0081, in part by the ZhuJiang (ZJ) Program of Guangdong under Grant 2023QN10X455, and in part by the Fundamental Research Funds for the Central Universities under Grant
2025ZYGXZR053. The associate editor coordinating the review of this article and approving it for publication was Dr. Long Chen. (Corresponding authors: Huan Wang; Qi Liu.)}
\thanks{Zhenyu Sun is with the School of Engineering, Westlake University, Hangzhou 310030, China, and also with the School of Future Technology, South China University of Technology, Guangzhou 511442, China (e-mail: ftsunzhenyu@mail.scut.edu.cn).}
\thanks{Xiaohan Zhang and Qi Liu are with the School of Future Technology, South China University of Technology, Guangzhou 511442, China (e-mail: ftxiaohanzhang@mail.scut.edu.cn; drliuqi@scut.edu.cn).}
\thanks{Huan Wang is with the School of Engineering, Westlake University, Hangzhou 310030, China (email: wanghuan@westlake.edu.cn).}
\thanks{\textcolor{softred}{$^{\dagger}$Corresponding author.}}}

\maketitle
\begin{abstract}
\label{sec:abstr}
\textcolor{softred} {Challenges remain in ego-centric 3D scene generation due to limited view overlap and the dominant influence of individual perspectives on scene interpretation.}
These factors hinder the creation of viewpoint-consistent and semantically aligned visual content, as well as the construction of accurate geometric structures.
In this paper, we propose \textbf{\textcolor{softred}{CGGS}}, a text-to-3D framework aiming to enhance 3D-content-awareness and address geometric distortions in ego-centric scene generation.
Firstly, the Ego-centric Generator is proposed by fine-tuning a Multi-View Latent Diffusion Model with consistency-augmented loss to generate consistent, high-fidelity 2D content aligned with textual descriptions.
Then, Layout Decorator leverages optical flow and point-track correspondence to estimate depth, therefore producing dense point clouds as coarse layouts from the ego-centric 2D priors.
Building on this initialization, Geometric Refiner is proposed to enhance 3D Gaussian reconstruction via an entropy-based Mutual Information Depth Loss (MID) combined with a hierarchical optimization scheme for improving visual quality and geometric structure.
Comprehensive experiments demonstrate that \textcolor{softred}{CGGS} outperforms previous methods in generating coherent and accurate text-driven 3D scenes.
Project page: \href{https://cggs-26.github.io/cggs26/}{\textcolor{softblue}{https://cggs-26.github.io/cggs26/}}.
\end{abstract}

\section{Introduction}
\label{sec:intro}
\begin{figure*}
    \centering
    \includegraphics[width=1\linewidth]{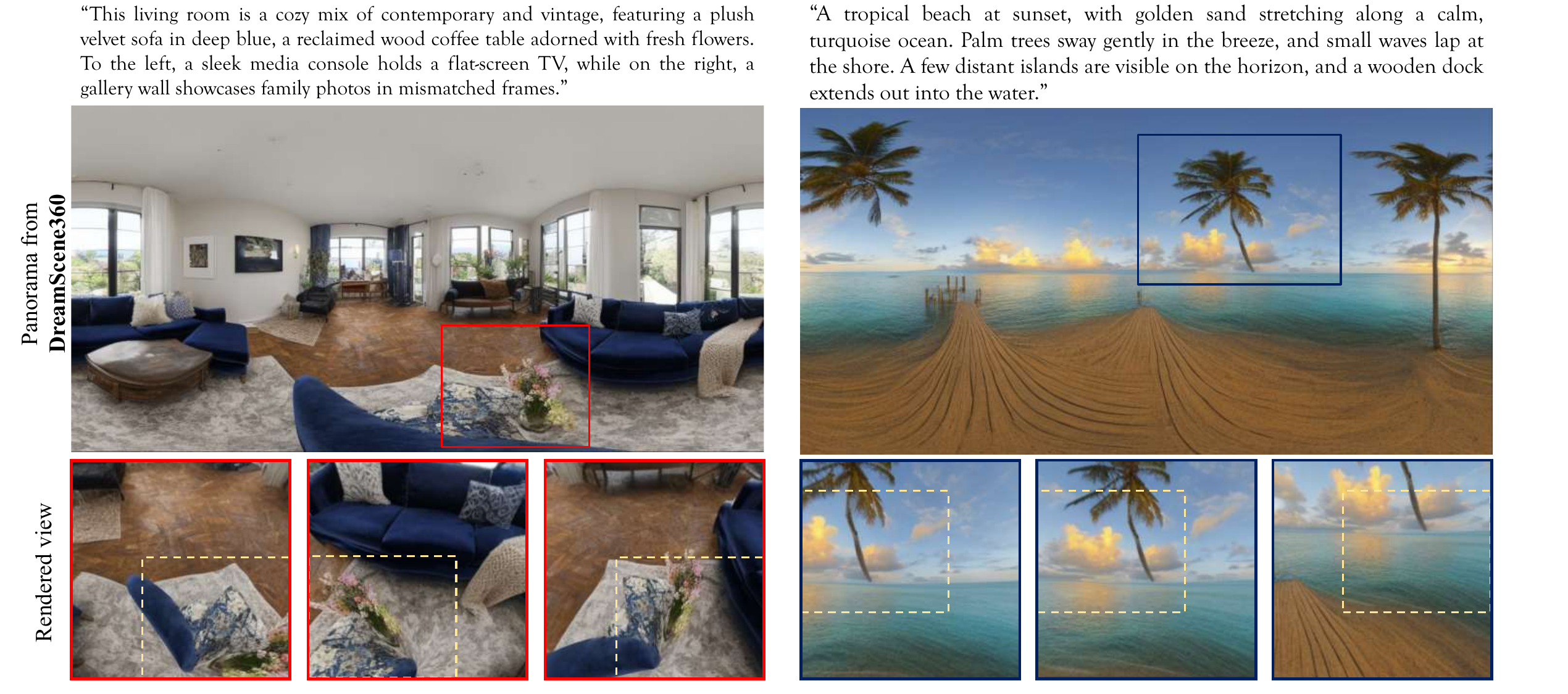}
    \caption{\textbf{Visualization of geometric distortions in panoramic generation} (using DreamScene360~\cite{zhou2024dreamscene360} as an example).
    \textbf{1) Insufficient Text-Content Alignment:} Significant textual details are omitted in the generation, such as the absence of "mismatched frames" on the gallery wall despite being explicitly specified in the prompt.
    \textbf{2) Polar Geometric Distortions:} Due to the inherent nature of equirectangular projection, severe radial stretching and bending occur near the top and bottom boundaries (e.g., the warped ceiling and distorted sand), which violates perspective consistency.
    \textbf{3) Unreasonable Structural Artifacts:} The model fails to maintain physical continuity, most notably evidenced by the severed and floating tree trunks in the beach scene, as well as incoherent horizon lines that hinder valid 3D reconstruction.
    }
    \label{fig:distortion_of_panorama}
\end{figure*}
3D scene generation has recently gained significant attention, fueled by advances in generative models and strong image priors. In particular, generating 3D scenes from textual descriptions holds great promise for a wide range of real-world applications in AR/VR, robotics, and autonomous driving. 
With the rapid development of text-to-image generation~\cite{nichol2021glide,ramesh2022hierarchical,saharia2022photorealistic}, progress has been made toward text-to-3D generation. 
Latent Diffusion Models (LDMs)~\cite{saharia2022photorealistic,rombach2022high} have been leveraged to optimize Neural Radiance Fields (NeRF)~\cite{mildenhall2020nerf} 
via CLIP embeddings~\cite{radford2021learning,jain2022zeroshottextguidedobjectgeneration} or score distillation sampling (SDS)~\cite{poole2023dreamfusion,lin2023magic3d,shi2024mvdream}.
However, these approaches often suffer from low rendering fidelity, multi-view inconsistencies, 
and limited scalability to scene-level 3D generation with fine-grained detail preservation.
In contrast, the progressive expansion from text-driven 2D priors to 3D content~\cite{hoellein2023text2room,chung2023luciddreamerdomainfreegeneration3d,zhang2024text2nerf,yu2024wonderjourney,shriram2025realmdreamer,zhou2025recurrentdiff} enables high-quality synthesis but accumulates errors across iterations that induce style inconsistencies 
and structural discrepancies.
Advance in 3D Gaussian splatting (3DGS)~\cite{kerbl2023gaussian} 
and feed-forward architectures has significantly enhanced 
generalizability and the fidelity of complex 3D content representations, 
catalyzing holistic and realistic text-to-3D scene generation under 
both forward-facing and center-convergent viewpoint settings~\cite{li2024scenedreamer360textdriven3dconsistentscene, 
zhou2024dreamscene360, zhou2024holodreamerholistic3dpanoramic, 
li2024director3d, yang2024prometheus}. 

\textcolor{softred}{Despite these efforts, ego-centric 3D scene generation faces a fundamental dilemma regarding the choice of 2D priors: panoramic versus multi-view representations.}
\textcolor{softred}{While panoramic generation naturally ensures global continuity with a unified $360^{\circ}$ field of view, the requisite equirectangular projection introduces severe geometric distortions—particularly near the poles—which fundamentally violate the pinhole camera assumption inherent in 3DGS and SfM pipelines.}
\textcolor{softred}{Such distortions inevitably lead to structural degradation and texture artifacts during the 3D lifting process, as illustrated in Fig.~\ref{fig:distortion_of_panorama}.}

\textcolor{softred}{Conversely, multi-view generation synthesizes perspective images that are geometrically distortion-free and rich in local high-frequency details, offering a mathematically robust foundation for high-fidelity reconstruction.}
\textcolor{softred}{However, this paradigm inherently struggles with inter-view consistency due to the lack of a unified canvas.}

\begin{figure*}[htbp]
    \centering    
    \includegraphics[width=\linewidth,trim=1.2cm 3.6cm 2.1cm 4.1cm, clip]{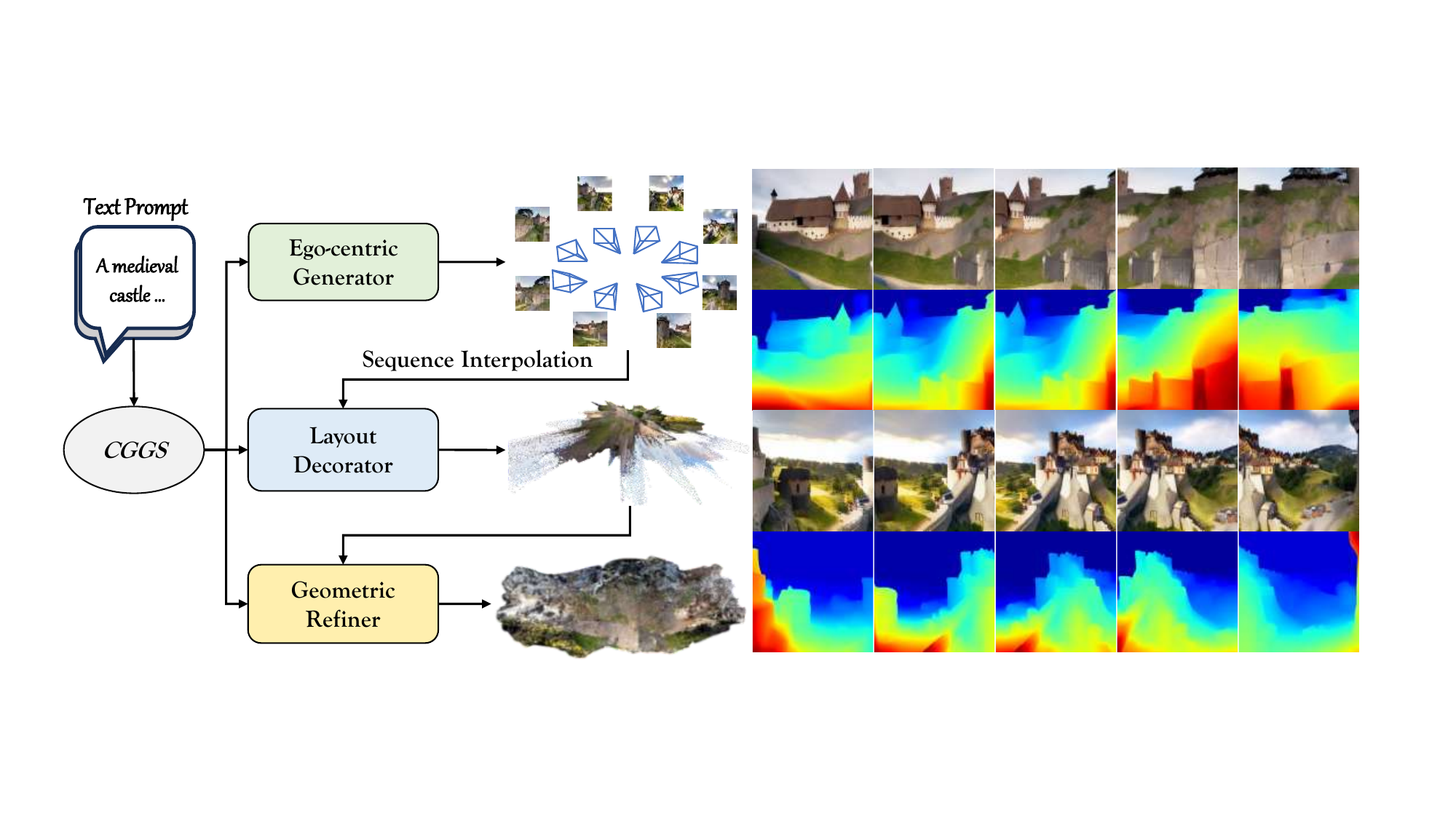}
    \caption{With text prompts as input, \textbf{\textcolor{softred}{CGGS}} employs three core components: the Ego-centric Generator creates ego-centric 2D priors, the Layout Decorator proposes additional scene details, and the Geometric Refiner further enhances the geometric structure and visual quality.}
\label{fig:teaser}
\end{figure*}

To tackle with the aforementioned issues, we introduce \textbf{\textcolor{softred}{CGGS}}, which unleashes the potential of latent diffusion models in text-image and image-image alignment, and learns the ego-centric 3D representation from the 2D images through a hierarchical 3D Gaussian optimization, as shown in Fig.~\ref{fig:teaser}.
We leverage the real-world datasets Matterport3D~\cite{Matterport3D},  RealEstate-10k~\cite{zhou2018stereo} and CO3Dv2~\cite{reizenstein21co3d} to achieve domain-free, realistic 3D generation from textual descriptions. 
Following the settings of Correspondence-Aware Attention (CAA) for multi-view generation~\cite{tang2023MVDiffusion}, we use Matterport3D~\cite{Matterport3D} to fine-tune our ego-centric multi-view generator from stable diffusion model~\cite{stabilityai2023stablediffusion}.
To enhance the semantic alignment and cross-view consistency, we introduce a consistency-augmented loss term as regularization to the LDM loss during the training of the CAA module.
Building upon the synthesized views, a Flow-Depth Estimator is used to generate a dense point cloud as layout initialization.
This approach can reconstruct a robust 3D structural layout of the scene from ego-centric 2D priors, whereas conventional Structure-from-Motion methods (SfM)~\cite{schonberger2016structure} typically struggle with such tasks.
Based on the initial 3D layout, we further leverage the Mutual Information Depth Loss (MID) to refine the scene during 3D Gaussian optimization, combined with a hierarchical optimization strategy, maintaining rendering robustness.
Collectively, our key contributions are as follows:
\noindent
\begin{itemize}
  \item Ego-centric Generator: A Multi-View Latent Diffusion Model is fine-tuned with our novel Consistency-Augmented Loss to produce ego-centric 2D priors that faithfully reflect the semantic intent of the textual descriptions and enhance cross-view consistency.
  \item Layout Decorator: A Flow-Depth Estimator guided by optical flow and point-track correspondences, transforming ego-centric 2D priors into a dense, coarse 3D layout. This approach addresses the inefficiencies and failure modes of direct SfM on ego-centric views.
  \item Geometric Refiner: Building upon the initial layouts, the hierarchical 3D Gaussian optimization supervised with Mutual Information Depth Loss (MID) iteratively sharpens structural details and enforces cross-view consistency, yielding geometrically precise and high-fidelity generation content.
\end{itemize}

\section{Related Work}
\label{sec:related_work}
\textbf{2D Contents Generation.} 
Generative Adversarial Networks (GANs)~\cite{goodfellow2014generative} were initially the leading method for image generation. 
Despite their success in creating 2D contents~\cite{viazovetskyi2020stylegan2,radford2016unsupervised,Isola_2017_CVPR,Karras_2019_CVPR}, GANs struggle with textual prompt interpretation and dataset-specific biases.
Diffusion models~\cite{song2019score,Ho20ddpm,song2020ddim,song2021score} have emerged as a promising alternative for image generation. 
They have built a strong foundation for customizing LDMs~\cite{ nichol2021glide,rombach2022high,saharia2022photorealistic} to produce domain-specific contents from textual descriptions.
Classifier-free guidance~\cite{ho2022classifier} is one such technique that has been employed to further enhance the fidelity to textual prompts.
Moreover, several works~\cite{berrada2024boosting,lin2024diffusion} further explore the application of perceptual loss on diffusion objectives to enhance image quality.
In recent times, several works~\cite{bar2023multidiffusion, li2023panogen, zhange2023diffcollage,wang2024customizing,panfusion2024} have achieved panorama generation from texts, but faced challenges in dealing with image distortions caused by projection methods.
In this study, we employ text-based multi-view generation to create ego-centric 2D priors, serving as systematic guidance for 3D scene generation from textual description. 

\textbf{Text-to-3D Generation.}
DreamFields~\cite{jain2022zeroshottextguidedobjectgeneration} pioneered the integration of vision-language models like CLIP~\cite{radford2021learning} with NeRF~\cite{mildenhall2020nerf} to synthesize 3D objects from textual descriptions, followed by~\cite{text2mesh2022} utilizing mesh as the 3D representation. 
Subsequent works employ 2D diffusion models to refine 3D representations through Score Distillation Sampling (SDS)~\cite{poole2023dreamfusion,lin2023magic3d,chen2023fantasia3d,shi2024mvdream} or Score Jacobian Chaining~\cite{wang2023score}.
ProlificDreamer~\cite{wang2023prolific} further optimized this approach by introducing Variational Score Distillation (VSD) to deal with the over-saturation problem.
Recent progress in text-to-image generation~\cite{nichol2021glide,ramesh2022hierarchical,stabilityai2023stablediffusion} with diffusion models has laid the groundwork for text-to-3D generation using LDMs~\cite{rombach2022high,saharia2022photorealistic,hong20243dtopia, zhou2025recurrentdiff}.
Nevertheless, most of these works mainly concentrate on object-level generation.

For scene-level 3D generation, various attempts~\cite{hoellein2023text2room,zhang2023rgbd2,ouyang2023text2immersiongenerativeimmersivescene,chung2023luciddreamerdomainfreegeneration3d,SceneScape,zhang2024text2nerf,wang2024vistadream} synthesize 3D scenes through progressive expansion by merging image inpainting models~\cite{rombach2022high} and monocular depth estimation models~\cite{ranftl2020towards,zoedepth,yang2024depth}.
Despite improvements for egocentric scenarios, these methods still lead to geometric and textural artifacts due to inherent inpainting limitations and depth‑alignment errors.
More recent works~\cite{yu2024wonderjourney,zhou2024dreamscene360,li2024scenedreamer360textdriven3dconsistentscene,zhou2024holodreamerholistic3dpanoramic} consider to develop 3D-scene generation from a panorama, but require extra multi-view constraints to reduce single-viewpoint limitations, thus producing suboptimal results of ego-centric scene representation.
Several works~\cite{li2024director3d,yang2024prometheus} propose end-to-end 3D generation frameworks, directly decoding 3D Gaussians from latent space. 
However, these approaches are restricted by relatively brief trajectory lengths and pose challenges in generating 3D models from purely outward-facing camera trajectories.
Therefore, developing 3D scene generation from ego-centric perspectives with cross-modal generalization remains a challenging issue.

\textbf{3D Scene Representation.} 
The development of 3D Gaussian Splatting (3DGS)~\cite{kerbl2023gaussian} has ushered in a new era of efficient, high-fidelity 3D reconstruction and synthesis~\cite{mildenhall2020nerf,you2023pointcloud,kerbl2023gaussian}, significantly reducing the rendering time compared to NeRF-based methods~\cite{mildenhall2020nerf,yu2021pixelnerf,bian2022nopenerf,metzer2023latentnerf,wynn2023diffusionnerf}. 
In particular, optimization over 3D Gaussians initialized from point clouds has emerged as a dominant paradigm.
These point clouds are most often obtained via Structure‐from‐Motion (SfM) pipelines~\cite{schonberger2016structure} or by back‐projecting pixels through monocular depth estimator (MDE)~\cite{ranftl2020towards,zoedepth,yang2024depth}. 
Several recent works \cite{pan2024glomap,smith2024flowmap} have advanced SfM by fusing global consistency constraints and dense geometric priors.
Moreover, targeting sparse-view reconstruction, various methods~\cite{tang2024lgm,zhang2024gslrm,xu2024grm,chen2024mvsplat,charatan2024pixelsplat,szymanowicz2024splatterimage} map pixel-aligned features into 3D Gaussians and optimize them end-to-end, albeit at the cost of substantial GPU resources and large-scale training data.
To further refine geometry, recent works~\cite{Li2024dngaussian,zhu2024fsgs,yu2024wonderworld} have introduced depth‐priors to enforce consistency between 3DGS parameters and geometric structure. 

Nevertheless, these works often focus on the object-centric situation, neglecting that ego-centric 3D reconstruction suffers from scarce cross-view overlap and pronounced viewpoint bias, hindering both semantic consistency and geometric accuracy.
Distinctively, our \textcolor{softred}{CGGS} first leverages an optical flow-guided depth estimator to construct a dense point cloud from ego-centric 2D priors, then refines this initialization with a Mutual Information Depth Loss (MID) and a hierarchical 3D Gaussian optimization to deliver view-consistent geometric representations.

\section{Preliminary}
\label{sec:preliminary}
\noindent \textbf{Latent Diffusion Models (LDMs)} \cite{saharia2022photorealistic,rombach2022high,nichol2021glide} operate generation in a learned latent space. 
They first train an auto-encoder~\cite{kingma2014autoencoding} to compresses high-dimensional data $x$ into a latent representation $z = \mathcal{E}(x)$, from which $x$ can be approximately reconstructed via $\hat{x} = \mathcal{D}(z)$.
Then a denoising network $\epsilon_\theta$ is employed to reverse a gradual noising process applied to the latents. 
The training loss minimizes the difference between added noise $\epsilon$ and predicted noise $\epsilon_\theta$:
\begin{equation}
\mathcal{L} = \mathbb{E}_{x,\epsilon \sim \mathcal{N}(0,1),t} \left[ \left\| \epsilon - \epsilon_\theta(z_t, c, t) \right\|_2^2 \right],
\label{equ:LDM}
\end{equation}
where the noised latent at timestep $t$ is represented as $z_t = \sqrt{\bar{\alpha}_t}\mathcal{E}(x) + \sqrt{1-\bar{\alpha}_t}\epsilon$, and $c$ denotes optional conditioning information. During generation, the model starts from random noise $z_T \sim \mathcal{N}(0,1)$, iteratively denoises it using $\epsilon_\theta$, and finally decodes the resulting latent into the output image.

\vspace{0.5em}
\noindent \textbf{Correspondence-Aware Attention (CAA).} MVDiffusion~\cite{tang2023MVDiffusion} introduces CAA blocks to ensure 3D-aware multi-view consistency.
The CAA block processes N feature maps concurrently.
For a source feature map \( F \), cross-attention is performed with \( (N-1) \) target feature maps \( F^l \). 
The message \( M \) is calculated as follows:
\begin{equation}
M = \sum_{l} \sum_{t^l \in \mathcal{N}} \mathrm{SoftMax}\left( W_Q \bar{F}(s) \cdot W_K \bar{F}^l(t^l_*) \right) W_V \bar{F}^l(t^l_*) \;,
\end{equation}

\begin{equation}
\bar{F}(s) = F(s) + \gamma(0) \;, \quad \bar{F}^l(t^l_*) = F^l(t^l_*) + \gamma(s^l_* - s) \;,
\end{equation}
where \( W_Q \), \( W_K \), and \( W_V \) are query, key, and value matrices. 
The position encoding \( \gamma(\cdot) \) is added to target features based on the 2D displacement between \( s^l_* \) and \( s \), which provides relative location within local neighborhoods. 
CAA blocks are integrated into the pre-trained stable diffusion UNet~\cite{stabilityai2023stablediffusion}, with other modules frozen during training to maintain original model functionality~\cite{tang2023MVDiffusion}. 
The following loss is used for training CAA blocks:
\begin{equation}
\begin{split}
\mathcal{L} := & \mathbb{E}_{\left\{\mathbf{Z}^i_t = \mathcal{E}(\mathbf{x}^i)\right\}_{i=1}^N, \left\{\boldsymbol{\epsilon}^i \sim \mathcal{N}(0, I)\right\}_{i=1}^N, \mathbf{y}, t} \\
& \Bigg[ \sum_{i=1}^N \left\| \boldsymbol{\epsilon}^i - \epsilon^i_{\theta}(\left\{\mathbf{Z}^i_t\right\}, t, \tau_{\theta}(\mathbf{y})) \right\|_2^2 \Bigg],
\end{split}
\label{equ:MVDiffusion loss}
\end{equation}
where $\epsilon_{i}$ and $\mathbf{Z}^i_t$  respectively denote the noise and the latent representation of the $i^{th}$ image $\mathbf{x}^i$, and $y$ represents the condition.
However, simply optimizing the KL divergence between the multi-view image forward process and the denoising process to align their distributions is insufficient for the model to balance multi-view consistency and text-image alignment.
Panfusion~\cite{panfusion2024} uses a dual-branch structure and adds a panoramic perspective to enhance text-image consistency.
Our approach, though, adopts a more elegant design without extra 2D architecture to leverage panoramic information.
\begin{figure*}[htbp]
    \centering    
    \includegraphics[width=\linewidth,trim=12.2cm 15.91cm 14.53cm 9.6cm, clip]{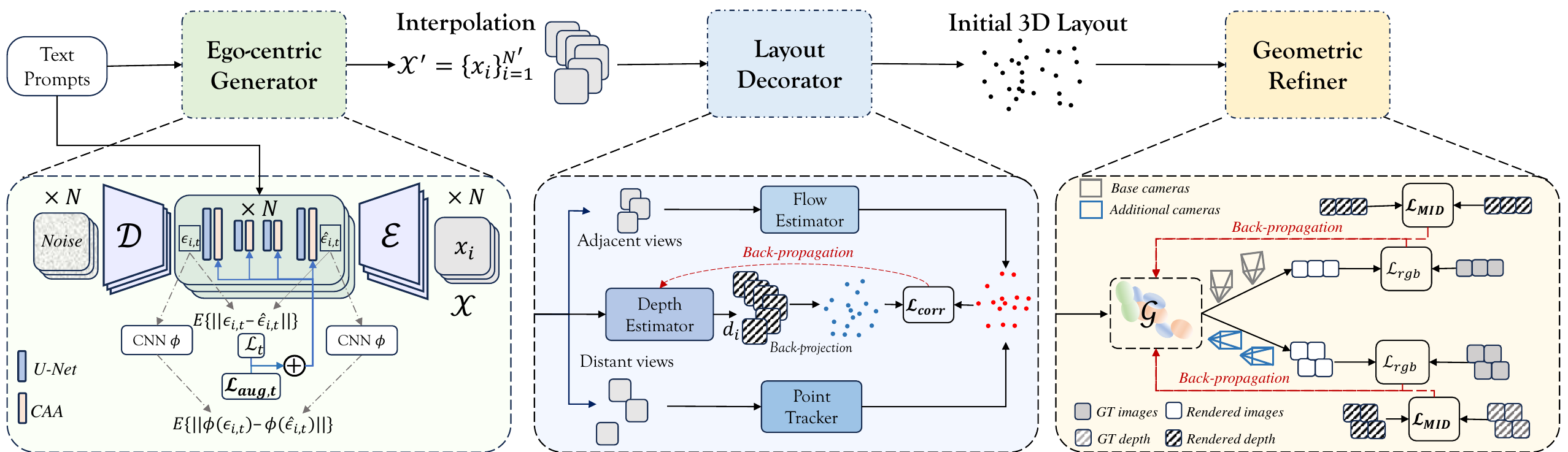}
    \caption{\textbf{Pipeline of CGGS}. It primarily comprises the Consistency-Augmented MV-LDM as Ego-centric Generator, the Flow-Depth Estimator as Layout Decorator, and the 3D Gaussian Optimization combined with MID Loss and the hierarchical optimization strategy, serving as Geometric Refiner.}
\label{fig:pipeline}
\end{figure*}

\section{Methodology}
\label{sec:methodology}
We set up our problem formulation as follows.
We consider the ego-centric multi-view images as $\mathcal{X} = \{x_i\}_{i=1}^N$, with $\mathcal{C} = \{c_i\}_{i=1}^N$ as the corresponding camera trajectory.  
The $\mathcal{C}$ can be described as known variable with different settings, and we consider each $x_i$ obeys the distribution $p\{x_i\,|\,x_1,\,...\,x_{i-1},\,x_{i+1},\,...\,,\,x_{N}, \mathcal{C}\}$.
The images are interpolated to $\mathcal{X}^{'}$ with $N^{\prime}$ views $(N^\prime > N)$, and we further model the ego-centric 2D priors as a motion sequence $\mathcal{M}$, aiming at deriving the dense point cloud as 3D layouts.
Additionally, each image $x_i$ is rendered from the 3D Gaussian representation $\mathcal{G}$ under camera pose $c_i$:  
$x_i = \mathcal{R}\bigl(\mathcal{G},\,c_i\bigr)$, where $\mathcal{R}$ denotes the rendering operator.
\textcolor{softred}{CGGS} addresses this task through three synergistic stages, as shown in Fig.~\ref{fig:pipeline}. 
Initially, a Multi-View Latent Diffusion Model (MV-LDM) is employed as the Ego-centric Generator to learn the conditional image distribution over specified camera trajectories and textual prompts, reinforced by a consistency augmentation loss as discussed in Sec.~\ref{subsec:ego-centric generator}. 
Directly conducting conventional SfM, e.g.~\cite{schonberger2016structure}, on ego-centric views yields inefficient and low-quality layouts. 
To address this, we introduce a Flow-Depth Estimator as Layout Decorator to generate a dense coarse point cloud, as elaborated in Sec.~\ref{subsec:layout decorator}. 
Finally, to achieve geometrically precise reconstructions, the 3D Gaussian optimization is refined with Mutual Information Depth loss and a hierarchical strategy, as presented in Sec.~\ref{subsec:geometric refiner}.

\subsection{Ego-centric Generator}
\label{subsec:ego-centric generator}
A crucial requirement of ego-centric generation is the holistic coherence and the semantic alignment.
The suboptimal text–content semantic alignment primarily arises from cross-view inconsistencies, including inconsistent representations of the same content across views and artifacts from fragmented structures or physically implausible structures across views.
We address these issues as follows.

Here we leverage the diffusion process from MVDiffusion~\cite{tang2023MVDiffusion} to simultaneously synthesize $N$ multi-view images  
that collectively capture a 360-degree scene coverage. 
However, during the training of multi-view generation, each view may introduce distinct gradient directions due to variations in perspective, leading to conflicting optimization signals.  
These discrepancies can hinder the effective minimization of the score matching loss, complicating the learning of a coherent representation across views.
Consequently, the capacity to accurately capture the underlying distribution diminishes, adversely affecting both inter-view consistency and alignment with the textual input.

To address the aforementioned issue, we propose a \textcolor{softred}{consistency-augmented-loss} on the latent diffusion objectives when training the CAA blocks, which boosts the alignment of gradients across perspectives.
Specifically, \textcolor{softred}{based on the Equ.~\eqref{equ:MVDiffusion loss}}, the score matching loss for each view $x_i$ is augmented with a regularization term defined as:  
\begin{equation}
\begin{split}
    \mathcal{L}_{\mathrm{aug}}
= & \mathbb{E}_{\left\{\mathbf{Z}^i_t = \mathcal{E}(\mathbf{x}^i)\right\}_{i=1}^N, \left\{\boldsymbol{\epsilon}^i \sim \mathcal{N}(0, I)\right\}_{i=1}^N, \mathbf{y}, t} \\
& \Bigg[\sum_{i=1}^N
\left\|
\phi\bigl(\epsilon^i\bigr)
-
\phi\bigl(\epsilon^i_\theta(\{\mathbf{Z}^i_t\},t,\tau_\theta(y))\bigr)\right\|_2^2\Bigg],
\label{equ:Loss aug}
\end{split}
\end{equation}
where $\phi$ denotes the multi-layer CNN $\phi$ with $L$ layers, whose hierarchical composition and parameter sharing impose a structured inductive bias that aligns per‑view gradient updates into a common subspace, thereby serving as a conflict harmonizer.
\textcolor{softred}{
In our framework, $\phi$ is instantiated as a VGG-16 network~\cite{simonyan2015very}.}
We utilize the full stack of convolutional
layers, therefore the $\mathcal{L}_{aug}$ integrates structural signals across multiple spatial scales in a unified manner.

Crucially, we employ He Initialization~\cite{HeZRS15} and keep the weights frozen without loading any pre-trained parameters.
This configuration transforms $\phi$ into a \textit{structured random projection}: the hierarchical convolutional layers impose a strong inductive bias for capturing spatial frequencies and geometry, while the random initialization ensures the feature space remains isotropic and free from the semantic prejudices (e.g., class-specific dominance) inherent in pre-training on ImageNet~\cite{deng2009imagenet}.
This yields a feature space that jointly exhibits (i) non-linear hierarchical structure, (ii) spatially localized multi-scale representation, and (iii) unbiased, stationary feature geometry—properties that are not simultaneously achieved by handcrafted filters or Fourier transform-based methods.
Because $\phi$ is identical and frozen across all views, it defines a stationary metric space for alignment.
Applying the chain rule to the augmented loss, the gradient with respect to the denoising parameters $\theta$ is derived as:
\begin{equation}\begin{split}\nabla_\theta \mathcal{L}_{\mathrm{aug}}
=& \mathbb{E}_{\left\{\mathbf{Z}^i_t = \mathcal{E}(\mathbf{x}^i)\right\}_{i=1}^N, \left\{\boldsymbol{\epsilon}^i \sim \mathcal{N}(0, I)\right\}_{i=1}^N, \mathbf{y}, t} \\
& \Biggl[\sum_{i=1}^N2 
\underbrace{(J_{\epsilon^i_\theta})^\top}_{\text{Backprop}}
\underbrace{(J_\phi)^\top}_{\text{Projection}}
\underbrace{\bigl(\phi(\epsilon^i_\theta)-\phi(\epsilon^i)\bigr)}_{\text{Feature Error}}
\Biggr],
\label{equ:gradient}
\end{split}
\end{equation}
\textcolor{softred}{where $J_\phi$ denotes the Jacobian of the harmonizer.}
\textcolor{softred}{Mathematically, the term $(J_\phi)^\top$ acts as a subspace projection operator.}
\textcolor{softred}{Unlike standard noise-prediction gradients, which are often high-frequency and mutually orthogonal across different ego-centric views (leading to optimization conflicts), Equ.~\eqref{equ:gradient} enforces the update vectors of all views to be projected onto the shared singular subspace of $J_\phi$.}
\textcolor{softred}{This alignment maximizes the cosine similarity between view-specific updates, yielding a consensus gradient that effectively counteracts incoherent noise while harmonizing the global geometric structure.
}

Thus, according to \textcolor{softred}{Equ.~\eqref{equ:MVDiffusion loss}} and Equ.~\eqref{equ:Loss aug}, the full training objective becomes:
\begin{equation}
    \mathcal{L}_{total}
=\mathcal{L} + \lambda_{aug}\mathcal{L}_{\mathrm{aug}}.
\label{equ:total ldm loss}
\end{equation}
This modification guarantees that each optimization step benefits from both precise score matching and a unified, multi-scale semantic prior that reduces inter-view gradient conflicts.
\begin{figure*}[htbp]
    \centering    
    \includegraphics[width=\linewidth,trim=4.75cm 4.0cm 7.15cm 3.6cm, clip]{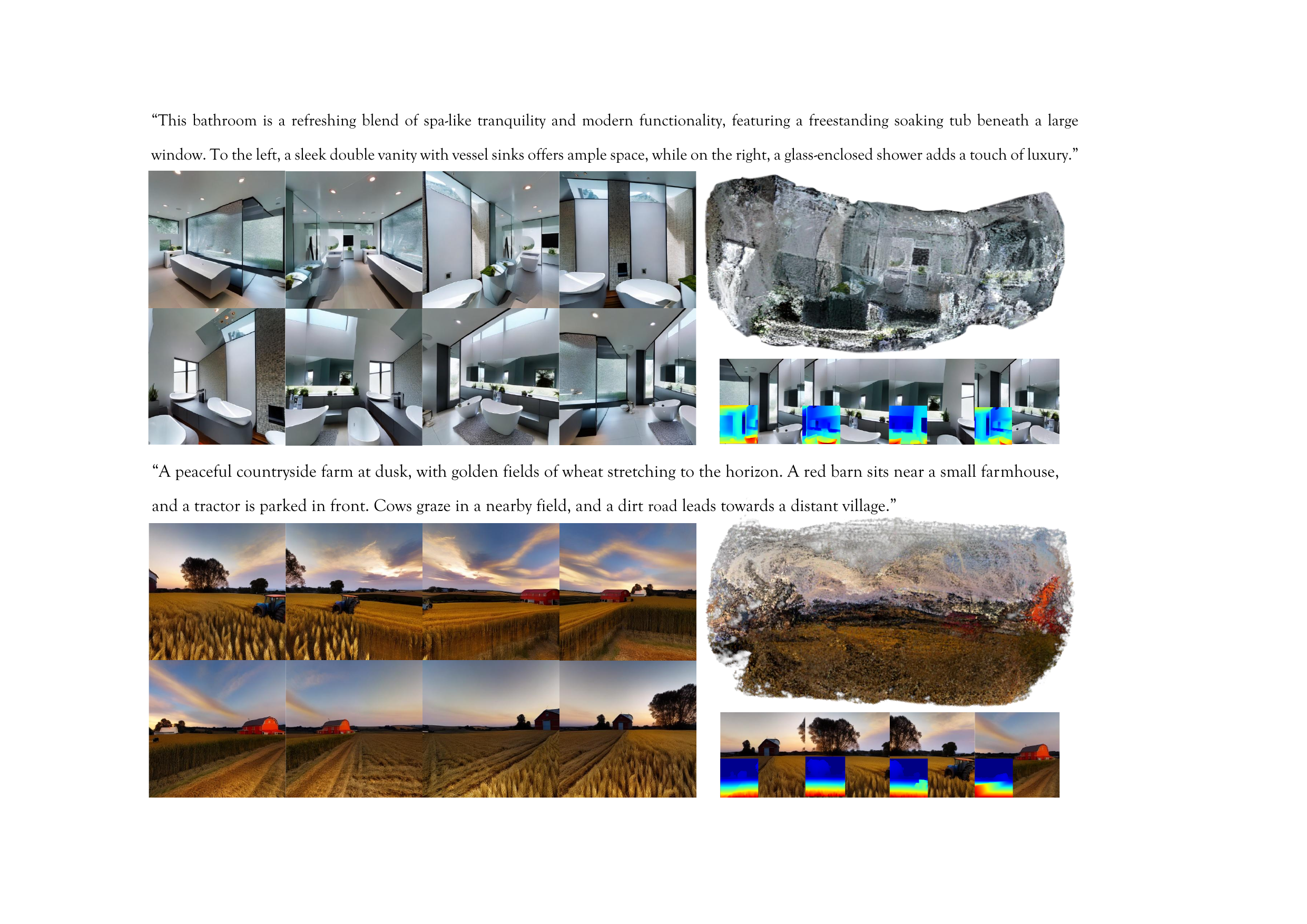}
    \caption{Generation results of \textcolor{softred}{CGGS} for ego-centric multi-view priors, gaussian point clouds, novel view synthesis, and depth maps. Our method generates harmonious, domain-free 3D scenes from ego-centric views, highly aligned with complex textual descriptions.}
\label{fig:generative results}
\end{figure*}

\subsection{Layout Decorator}
\label{subsec:layout decorator}
\textcolor{softred}{
While contemporary off-the-shelf monocular depth estimators~\cite{ranftl2020towards, zoedepth, yang2024depth} achieve impressive per-pixel accuracy, they inherently suffer from scale ambiguity and inconsistent geometric deformations across independent ego-centric views. 
Direct back-projection from such inconsistent priors results in stratified and misaligned point clouds.
To resolve this, we leverage flow and point correspondences as relative geometric constraints to enforce cross-view alignment.
}
To fully exploit the 2D priors, we first interpolate the image sequences and then upsample them according to predefined camera groupings.
This step aims to enhance the image resolution and provide more detailed information.
Then we treat the input image sequence $\mathcal{X^\prime}=\{x_i\}_{i=1}^{N^\prime}$ as a short video stream and model the motion sequences as $\mathcal{M} = \{\mathcal{X^\prime},\mathcal{F}\}$.
Here $\mathcal{F}$ represents the  sequence of optical flow from off-the-shelf flow estimator~\cite{teed2020raft}, which is defined as:
\textcolor{softred}{
\begin{equation}
    \mathcal{F} = \{f_i: x_i \leftrightarrow x_{i+1}\}_{i=1}^{N^{'}-1} \cup \{f_{N^{'}}: x_{N^{'}} \leftrightarrow x_1\}.
\end{equation}}
This establishes dense pixel‐level correspondences across adjacent ego‐centric views.
Theoretically, optical flow can also be used to directly predict pixel correspondences across views:
\begin{equation}
p_t = p_s + f_{st}(p_s),
\label{equ:flow correspondence}
\end{equation}
where $p_t$ denotes the target pixels in image$j$, and $p_s$ represents the source pixels in image$i$.
When $j = i+1$ for $i = 1,\dots, N'-1$ and $j = 1$ for $i = N'$, it corresponds to the situation between neighboring viewpoints. 

\textcolor{softred}{
Relying solely on short-term flow can lead to cumulative drift. 
To mitigate this, we integrate long-term Point Tracks~\cite{karaev2024cotracker} to rectify flow divergence over extended temporal windows explicitly. 
Crucially, the depth estimation network functions as a geometric regularizer during this supervised process; it filters out high-frequency noise from the flow estimators while learning the dominant, consistent structural trends mandated by the point tracks. 
Thus, rather than propagating error, this joint optimization effectively harmonizes the scale and geometry of the initial depth priors, yielding the consistent per-frame depth maps $\mathcal{D}=\{d_{i}\}_{i=1}^{N}$.}

Using the estimated depth $d_i$ and the known intrinsic calibration $K$, the pixel $p_s$ in the $i_th$ view is back-projected into 3D space as $z_s\in\mathbb{R}^3$.
Given the relative extrinsic transformation between the $i^{th}$ and $j^{th}$ cameras, the corresponding pixel coordinate $p_{t}$ of $p_{s}$ in the $j^{th}$ image is obtained by:
\begin{equation}
    \tilde{p_t}\,=\,K\,T_j\,T_i^{-1}\,
    \begin{pmatrix}
      z_s \\[2pt]
      1
    \end{pmatrix},
\label{equ:projection transform}
\end{equation}
where $T_i$ and $T_j$ are the $4\times4$ extrinsic matrices for views $i$ and $j$, respectively.
After dividing by its third component, the final pixel location $\hat{p}_t$ is produced.
According to the results from Equ.~\eqref{equ:flow correspondence}, a correspondence loss can be defined as: 
\begin{equation}
    \mathcal{L}_{corr}=\|\hat{p}_{t}-p_{t}\|.
\end{equation}
Minimizing this loss yields a supervised signal that iteratively enhances the depth estimation network, improving 3D layouts.
This formulation jointly models the 3D layout from ego-centric observations, effectively addressing the limitations of traditional SfM pipelines~\cite{schonberger2016structure} and mitigating the spatial misalignment issues commonly encountered in monocular depth estimation. 
Building upon the estimated depth maps $\mathcal{D}=\{d_i\}_{i=1}^{N'}$, the pixels are back-projected and merged into a unified coarse point cloud, yielding a consistent initial 3D scaffold that faithfully reflects the underlying scene geometry despite narrow baselines and frequent occlusions. 

\begin{figure*}[htbp]
    \centering    
    \includegraphics[width=\linewidth,trim=1cm 0.5cm 12cm 0.0cm, clip]{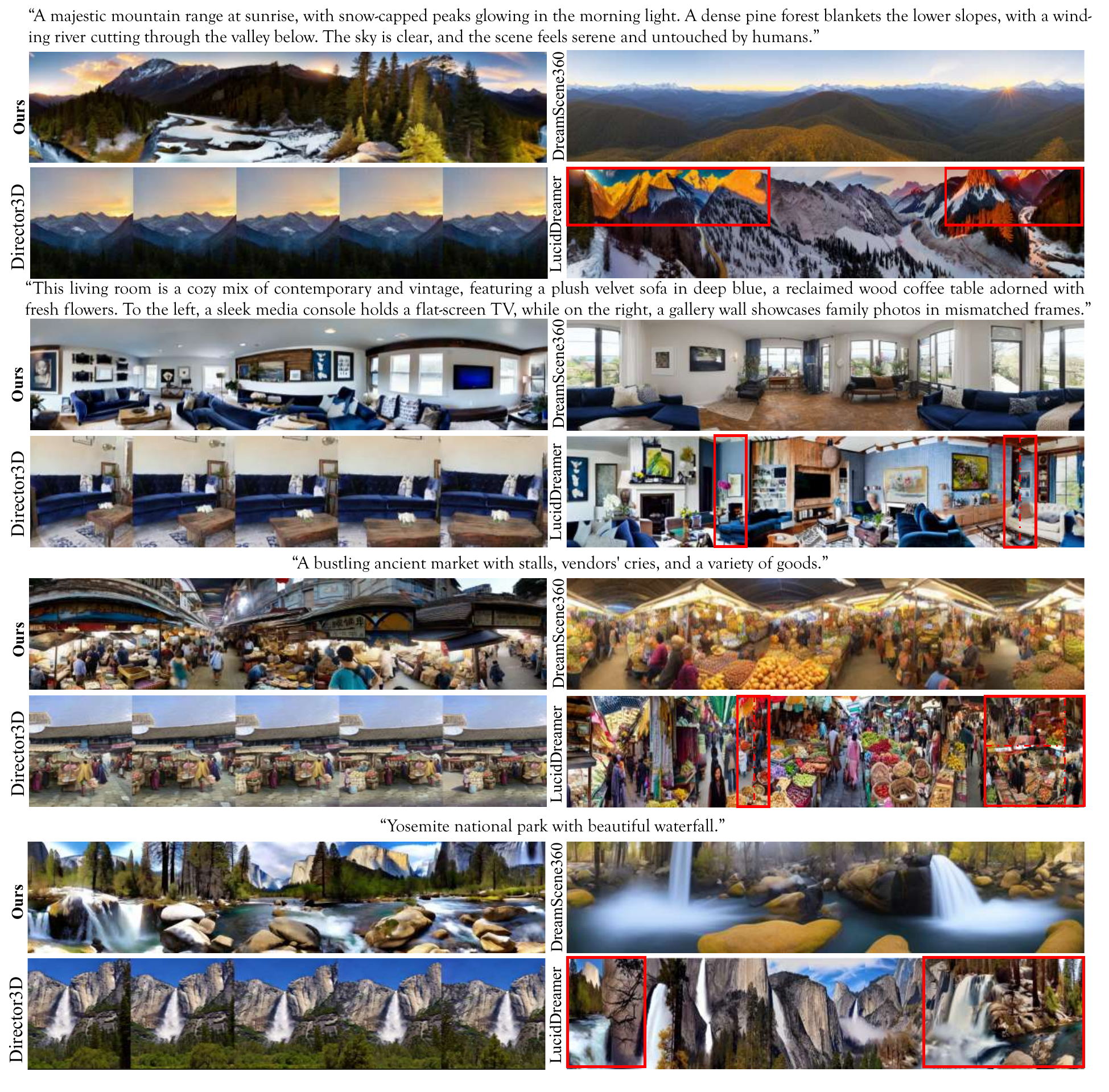}
    \caption{Qualitative comparison between \textcolor{softred}{CGGS} with other baselines. Our \textcolor{softred}{CGGS} produces multi-view images with rich detail and superior semantic coherence, showcasing domain‑agnosticity. Our results outperform other works with an accurately detailed description and unified 3D consistency. \textcolor{softred}{Specifically, DreamScene360 generates visual results with less major content in the horizon field; While Director3D is capable of depicting the content described in text prompts, it is constrained by a limited field of view; LucidDreamer causes undesirable style transfer, wrong stitches between concepts, and inconsistent content, as highlighted in the red box.}}
\label{fig:qualitative comparison}
\end{figure*}
\subsection{Geometric Refiner}
\label{subsec:geometric refiner}
Based on the point clouds and structural information generated by the Layout Decorator, Geometric Refiner further incorporates depth-aware geometric structure into the original 3DGS framework~\cite{kerbl2023gaussian}. 
This guides the cloning and splitting of Gaussian ellipsoids, enabling finer detail representation while preserving the overall geometric consistency. 
Previous works have introduced depth-based regularization to improve geometric accuracy, such as applying epipolar constraints~\cite{zheng2025nexusgs} or Pearson correlation-based depth supervision~\cite{zhou2024dreamscene360}. 
To enhance the ability to capture complex depth relationships and to mitigate its tendency toward oversmoothing, we further model the geometric supervision process using a mutual information depth loss, which provides a more expressive and robust signal for learning fine-grained depth structures. 
Specifically, for the perspective of the $i^{th}$ camera, the depth map $d_i$ from the Layout Decorator is considered as the ground truth depth, with its rendered depth map $d^{i}_{render}$ calculated from the differential rasterization of 3DGS.
We begin by flattening the rendered depth map $d^i_{\mathrm{render}}$ and the ground-truth depth map $d_i$ into one-dimensional vectors $r$ and $g$, respectively, which we treat as samples of continuous random variables $D_{\mathrm{render}}$ and $D_{\mathrm{gt}}$. The mutual information between these variables is defined as:
\begin{equation}
\mathcal{I}\bigl(D_{\mathrm{render}};D_{\mathrm{gt}}\bigr)
= \sum\sum p(r,g)\,
  \log\!\frac{p(r,g)}{p(r)\,p(g)}\,,
\label{mutual information}
\end{equation}
where $p(r,g)$ denotes the joint probability density of $r$ and $g$, and $p(r)$, $p(g)$ are their marginal densities.
Based on the operation, the MID loss is expressed as:
\begin{equation}
    \mathcal{L}_{\mathrm{MID}}
= 1 - \mathcal{I}\bigl(D_{\mathrm{render}};D_{\mathrm{gt}}\bigr)\,.
\label{equ:MID loss}
\end{equation}
Therefore, the total training loss is shown as follows:
\begin{equation}
\mathcal{L}^{rec}_{total}\,=\,(1-\lambda_{\mathrm{MID}})\mathcal{L}_{rgb}\,+\,\lambda_{\mathrm{MID}}\,\mathcal{L}_{\mathrm{MID}},
\label{equ:MID}
\end{equation}
where $\lambda_{\mathrm{MID}}$ is a scalar weight balancing this term against RGB objectives in the overall training loss.

\textcolor{softred}{Crucially, distinct from the common scale-invariant depth loss (e.g., Pearson correlation), which implicitly assumes a linear relationship between predicted and reference depths, the proposed MID loss operates on the statistical dependency between distributions. While Pearson-based objectives effectively handle global-scale ambiguity, they are prone to suppressing high-frequency geometric details and are sensitive to outliers, often resulting in over-smoothed structures. In contrast, by maximizing the mutual information, $\mathcal{L}_{\text{MID}}$ enforces a stricter alignment of the underlying structural entropy. This enables the optimization to capture complex, non-linear geometric correspondences and preserve sharp depth discontinuities, thereby yielding higher fidelity in ego-centric 3D reconstruction.}

During the Gaussian optimization stage, Geometric Refiner introduces supplementary cameras to mitigate the bias arising from a limited set of viewpoints. 
Inspired by the spatial supplementary camera arrangement in HoloDreamer~\cite{zhou2024holodreamerholistic3dpanoramic} and the multi-stage virtual camera strategy of DreamScene360~\cite{zhou2024dreamscene360}, we progressively expand the base cameras with additional cameras sharing intrinsic parameters, providing diverse viewpoints from different hierarchies. 
As training proceeds, both the number of supplementary cameras and their relative pose offsets increase, depicted as:
\begin{equation}
    \begin{aligned}
\mathcal{C}_{k} &= \mathcal{C}_{k-1} \;\cup\;
   \bigl\{\,\bigl(R_i\,\Delta R_{k},\,t_i + \Delta t_{k}\bigr)\bigr\},
&&\text{for }k=1,2,\dots,n \\
\mathcal{C}_0 &= \{\,(R_i,\,t_i)\}_{i=1}^{N^{'}}, 
&&\text{(base cameras)}\\
\end{aligned}
\end{equation}
where each $\Delta R_{k}$ and $\Delta t_{k}$ denotes the rotational and translational offsets at stage $k$, and $n$ is the total number of the hierarchical expansion stages. 
The loss at stage $k$ is computed by rendering from all cameras in $\mathcal{C}_{k}$ and comparing against the depth and RGB supervision signals.
This gradual expansion encourages reconstruction robustness and enhances the integrity of the 3D Gaussian optimization.

\section{Experiments}
\label{sec:experiments}

\subsection{Implementation Details}
\label{subsec:implementation details}
\textcolor{softred}{CGGS} leverages Matterport3D~\cite{Matterport3D} datasets, RealEstate10k~\cite{zhou2018stereo} and CO3Dv2~\cite{reizenstein21co3d} as real-world multi-view datasets.
The text prompts for scene description imitate the sentence structure used in MVDiffusion~\cite{tang2023MVDiffusion} and are expanded using GPT-4~\cite{openai2024gpt4} across various scenarios.
The number of views $N$ in the Ego-centric Generator is set to 8, with a horizontal field of view (FOV) $\Theta=90^{\circ}$ and a rotation angle $\theta=45^{\circ}$, and the $\lambda_{aug}$ is assigned to $0.5$.
For Layout Decorator, the interpolation number $N^{\prime}=20$, with FOV $\Theta^{\prime}=60^{\circ}$.
We set the $\lambda_{\mathrm{MID}}$ to $0.05$ and the hierarchical expansion epochs $n$ to $3$.
More details are reported below.

\vspace{0.3em}
\textbf{Datasets.} For multi-view generation, we leverage Matterport3D~\cite{Matterport3D} to enable text-driven ego-centric generation. 
For real-world multi-view datasets, we leverage RealEstate-10k~\cite{zhou2018stereo} and CO3Dv2~\cite{reizenstein21co3d} for accurate flow depth estimation at both scene and object levels.
Matterport3D is a large-scale indoor RGB-D dataset comprising 194,400 images and 10,800 panoramic views across 90 building-scale scenes.
RealEstate-10k consists of 10 million frames extracted from approximately 80,000 video clips sourced from about 10,000 YouTube videos.
CO3Dv2 is a collection of 1.5 million frames from 18,619 videos, covering 50 MS-COCO object categories for 3D reconstruction tasks.

\vspace{0.3em}
\textbf{Ego-centric Generator.} We fine-tune the CAA blocks with the improved loss function, as shown in Equ.~\eqref {equ:total ldm loss}. 
It takes about $35\sim40$ hours on four NVIDIA RTX A6000 GPUs, with a batch size equal to $4$ on each GPU. 
Ego-centric Generator generates $8$ ego-centric multi-view images with a resolution of $512\,\times\,512$.
For the sequence interpolation mentioned in Sec.~\ref{subsec:layout decorator}, we first fuse the image sequence $\mathcal{X}$ into a pseudo-panorama with the resolution of $4096\,\times\,968$, then project the view to acquire $20$ multi-view images with the size as $512\,\times\,512$.

\textbf{Layout Decorator.} The Flow-Depth Estimator is trained with RealEstate-10k and CO3Dv2 on one single NVIDIA RTX A6000 GPU for about 25 hours.
During the inference stage, it takes about 10 minutes to lift a specific scene as a dense point cloud from the multi-view priors.

\begin{table*}[htbp]
\centering
\caption{Quantitative comparison between \textcolor{softred}{CGGS} and other baselines. We benchmark our methods with other brilliant prior works across 24 scenes, covering indoor and outdoor environments. We report the evaluation in metrics that reflect both generation quality and reconstruction quality. The results indicate that \textcolor{softred}{CGGS} achieves the best overall performance, generating a semantically consistent 3D scene with high visual quality and proper geometric structure.}
\label{tab:quantitative comparison}

\resizebox{\textwidth}{!}{
\setlength{\tabcolsep}{4pt}
\begin{tabular}{@{}l c ccccc ccc@{}}
\toprule
\multirow{2}[2]{*}{Method} & \multirow{2}[2]{*}{3D Representations} & \multicolumn{5}{c}{Generation Quality} & \multicolumn{3}{c}{Reconstruction Quality} \\
\cmidrule(lr){3-7} \cmidrule(l){8-10}
 &  & CLIP Score $\uparrow$ & Sharp $\uparrow$ & Color $\uparrow$ & Resolution $\uparrow$ & Q-Align $\uparrow$ & PSNR $\uparrow$ & SSIM $\uparrow$ & LPIPS $\downarrow$ \\
\midrule
Text2Room~\cite{hoellein2023text2room} & Mesh & 24.732 & 0.215 & 0.210 & 0.231 & 0.697 & \textcolor{softred}{20.915} & \textcolor{softred}{0.844} & \textcolor{softred}{0.169} \\
LucidDreamer~\cite{chung2023luciddreamerdomainfreegeneration3d} & \textcolor{softred}{3DGS} & \underline{25.736} & 0.216 & \underline{0.211} & 0.224 & 0.764 & 25.667 & 0.824 & 0.163 \\
Director3D~\cite{li2024director3d} & 3DGS & 24.996 & \textbf{0.221} & \textbf{0.225} & \underline{0.232} & 0.754 & - & - & -\\
DreamScene360~\cite{zhou2024dreamscene360} & 3DGS & 25.022 & \underline{0.219} & 0.204 & \textbf{0.239} & \underline{0.828} & 32.587 & 0.969 & \underline{0.0477} \\
\midrule
\textbf{\textcolor{softred}{CGGS}} & 3DGS & \textbf{26.253} & 0.218 & \underline{0.211} & 0.231 & \textbf{0.839} & \textbf{37.345} & \textbf{0.977} & \textbf{0.0193} \\
\bottomrule
\end{tabular}
}
\end{table*}

\vspace{0.3em}
\textbf{Geometric Decorator.} The hierarchical optimization strategy is implemented as a staged process. During the warm-up phase, the 3DGS optimization proceeds with the default configuration. 
At a predefined initialization iteration, a set of $m$ additional cameras is introduced, sharing the same intrinsic parameters $\mathbf{K}$ and extrinsic parameters $\mathbf{E_{ij}}$, and arranged around each base camera with extrinsics $\mathbf{E_i}$, where $j = 1, 2, \ldots, m$. 
At each subsequent stage $k$, four additional cameras are further deployed around each base camera. 
As described in Sec.~\ref{subsec:geometric refiner}, the transformations $\Delta R_k$ and $\Delta t_k$ represent the relative pose offsets between each auxiliary camera $C_{ij}$ at stage $k$ and its corresponding base camera $C_i$. 
These offsets are progressively increased across stages to introduce more significant viewpoint variation, thereby enhancing structural representation and optimizing scene reconstruction.
The total number of hierarchical stages $n$ is set to be $3$, as mentioned in Sec.~\ref{subsec:geometric refiner}.

In Sec.~\ref{subsec:geometric refiner}, the theoretical training loss in Equ.~\eqref{equ:MID} for 3D Gaussian optimization can be expressed as
\begin{equation}
\mathcal{L}^{rec}_{total}\,=\,(1-\lambda_{\mathrm{MID}})\mathcal{L}_{rgb}\,+\,\lambda_{\mathrm{MID}}\,\mathcal{L}_{\mathrm{MID}},
\label{equ:appendix_MID}
\end{equation}
and the $\lambda_{\mathrm{MID}}$ is set to be $0.05$, as mentioned in Sec.~\ref{subsec:implementation details}.
Specifically, the detailed reconstruction training loss can be represented as:
\begin{equation}
\begin{split}
\mathcal{L}^{rec}_{total}\,= &\,\lambda_{SSIM}(1-SSIM)\,\;+\,\lambda_{\mathrm{MID}}\,\mathcal{L}_{\mathrm{MID}}\\
&+\,(1-\lambda_{SSIM}-\lambda_{MID})\,L_1\,,
\label{equ:appendix_MID_detailed}
\end{split}
\end{equation}
where $L_1$ denotes the per‑pixel absolute difference between the reference image and the rendered image. 
Here $\lambda_{SSIM}$ is set to be $0.2$ in practice.
We disable the opacity reset process to accelerate convergence and maintain high rendering quality.
The remaining configurations are consistent with 3DGS~\cite{kerbl2023gaussian}.
It takes about 3 minutes per scene for optimization.

\subsection{Generation Results}
\label{subsec:generation results}
Our generation results are presented in Fig.~\ref{fig:generative results}.
The ego-centric 2D priors exhibit strong cross-view consistency in both style and content and are well aligned with text prompts, reflecting the effectiveness of the Ego-centric Generator.
Moreover, our method supports vivid 3D scene generation from complex prompts, resulting in geometrically coherent structures that highlight the effectiveness of the Layout Decorator and Geometric Refiner.

\subsection{Qualitative Comparison}
\label{subsec:qualitative comparison}
We conduct qualitative comparisons between our method and several baselines using 3DGS~\cite{kerbl2023gaussian} as scene representation: LucidDreamer~\cite{chung2023luciddreamerdomainfreegeneration3d}, Director3D~\cite{li2024director3d}, and DreamScene360~\cite{zhou2024dreamscene360}.
As shown in Fig.~\ref{fig:qualitative comparison}, \textcolor{softred}{CGGS} generates 3D scenes with abundant geometric details and produces high-fidelity novel views.
LucidDreamer exhibits abrupt shifts in overall style and content elements, even under modest viewpoint variation.
Director3D tends to produce simple layouts when faced with intricately detailed text, resulting in visuals of suboptimal quality.
DreamScene360 demonstrates strong panoramic consistency at the global level but struggles to reproduce fine structural details and fully reflect the textual descriptions within its horizontal views.
These comparisons indicate that our \textcolor{softred}{CGGS} outperforms other methods in generating realistic 3D scenes with proper geometric details.

\subsection{Quantitative Comparison}
\label{subsec:quantitative comparison}
We tabulate relative metrics of generation quality and reconstruction quality to assess our \textcolor{softred}{CGGS} with other baselines~\cite{hoellein2023text2room,chung2023luciddreamerdomainfreegeneration3d,li2024director3d,zhou2024dreamscene360}.
For evaluation purposes, we employ GPT-4~\cite{openai2024gpt4} to generate a diverse set of text descriptions from multiple types of scenes from public datasets~\cite{Knapitsch2017Tanks,tang2023MVDiffusion,Dai2017ScanNet}, varying in complexity, style, and semantic content.
(1) Generation Quality: we adopt CLIP-Score~\cite{hessel2021clipscore} to assess the semantic alignment.
Q-Align~\cite{wu2024qalign} and CLIP-IQA~\cite{wang2023exploring}, including Sharp, Colorful, and Resolution, are used to measure image perceptual quality.
(2) Reconstruction Quality: for assessment of rendering quality, we adopt PSNR, SSIM, and LPIPS as metrics commonly used in 3D reconstruction tasks.
CLIP-Score (CS) is computed between the textual descriptions of each scene and the corresponding generated multi-view images, and the average is reported.
Q-Align and CLIP-IQA, which include Sharp (SH), Colorful (CL), and Resolution (RS), are evaluated on the multi-view images generated for each scene, with scores averaged per scene.
PSNR, SSIM, and LPIPS are used to assess view rendering quality in methods that generate scenes with reference images.

\begin{figure}
    \centering
    \includegraphics[width=1\linewidth, trim=3.2cm 8.5cm 8.8cm 2cm, clip ]{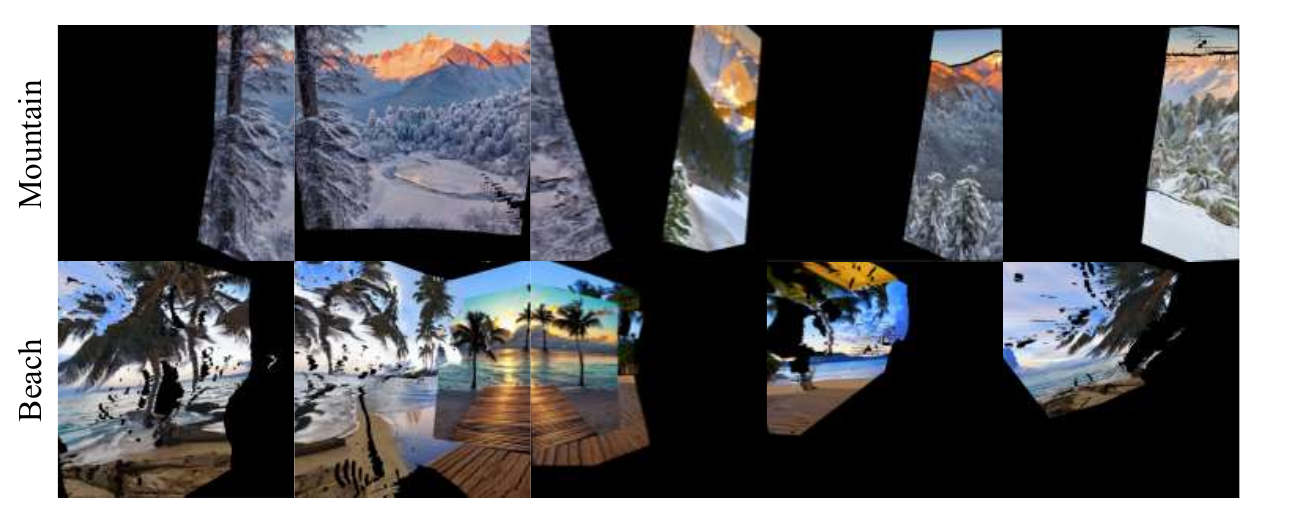}
    \caption{\textcolor{softred}{\textbf{Poor quality of the rendered views from Text2Room~\cite{hoellein2023text2room} on outdoor scenes}. There are large black artifacts and missing geometry.}}
    \label{fig:text2room bad performance}
\end{figure}

\textcolor{softred}{For Text2Room, since its rendered views contain large black artifacts and missing geometry as shown in Fig.~\ref{fig:text2room bad performance}, we only calculated its performance metrics on indoor scenes in Tab.~\ref{tab:quantitative comparison}.}
For LucidDreamer, which requires both a textual description and a single initial image as input, we use the first image from the multi-view sequence generated by \textcolor{softred}{CGGS} as its initial input.
As for Director3D, since it does not utilize any intermediate reference images, we do not evaluate its image reconstruction quality.

\textcolor{softred}{According to Tab.~\ref{tab:quantitative comparison}, our CGGS method demonstrates a comprehensive advantage over existing methods across multiple dimensions.
In terms of generation quality, CGGS achieves the highest CLIP Score (26.253), indicating superior semantic alignment between the generated 3D scenes and textual descriptions. 
While competitive methods like Director3D and DreamScene360 exhibit strengths in local attributes such as sharpness and resolution, CGGS delivers the best overall perceptual performance, as evidenced by its leading Q-Align score (0.839).
Simultaneously, CGGS exhibits exceptional performance in reconstruction fidelity, delivering a PSNR of 37.345 and an LPIPS of 0.0193, which underscores its high structural accuracy.
Text2Room and LucidDreamer suffer from incongruent stitching artifacts, where unrelated semantic concepts are incoherently merged. 
This leads to degraded rendering quality, particularly evident in suboptimal LPIPS scores.
These results demonstrate the robustness and effectiveness of \textcolor{softred}{CGGS} in generating semantically consistent and visually high-fidelity 3D scenes.}

\subsection{Ablation Study}
\label{subsec:ablation study}
\begin{figure*}
    \centering
    \includegraphics[width=1\linewidth, trim=2.3cm 9.5cm 5.8cm 0cm, clip ]{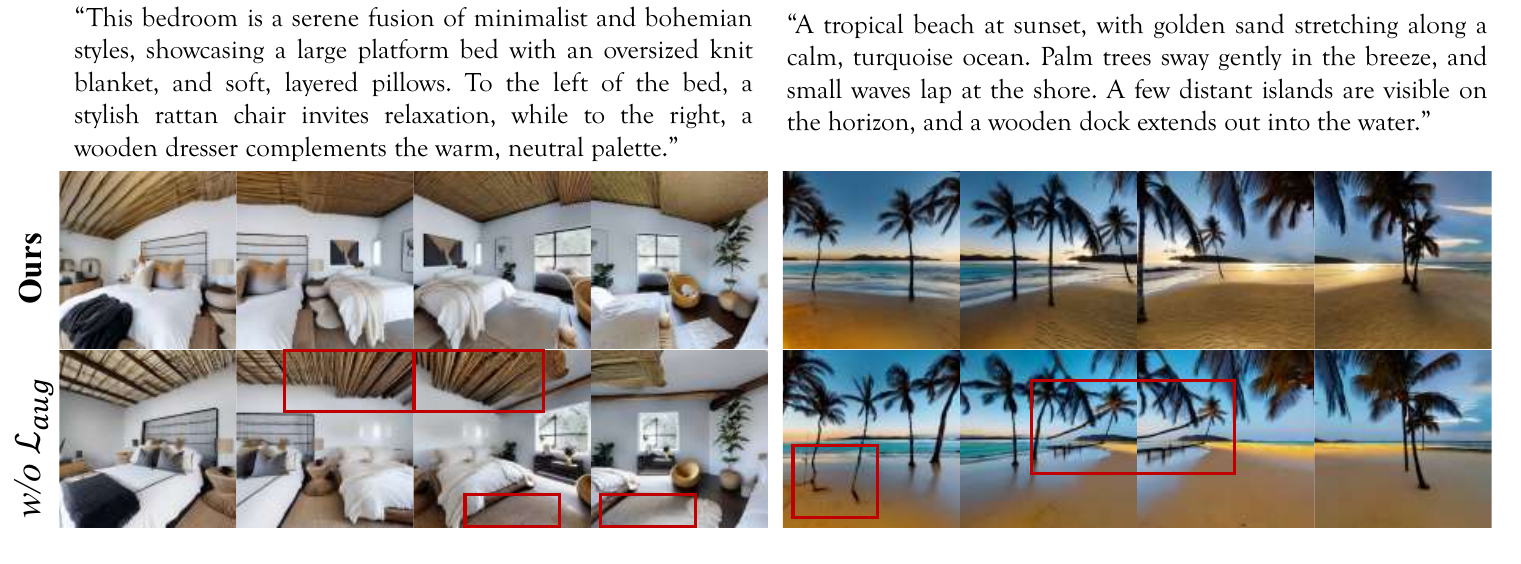}
    \caption{\textbf{Ablation study on consistency-augmented loss $\mathcal{L}_{aug}$.} Without $\mathcal{L}_{aug}$, cross‑view texture discrepancies become pronounced, with abrupt background artifacts (e.g., exposed ceilings in bedroom scenes) and physically implausible anomalies (e.g., floating, distorted trees on beaches) emerging.}
    \label{fig:ablation_study_NAMV_LDM}
\end{figure*}

\vspace{0.3em}
\textbf{Ego-centric Generator.} In the architecture of \textcolor{softred}{CGGS}, the Ego-centric Generator is crucial for enhancing semantic alignment and cross-view coherence.
With $\mathcal{L}_{aug}$ removed, as illustrated in Fig.~\ref{fig:ablation_study_NAMV_LDM}, the content layout becomes chaotic, cross‐view consistency of texture details degrades, and implausible image artifacts emerge.
\begin{table*}[htbp]
  \centering
  \caption{Ablation studies of Ego-centric Generator on consistency-augmented loss $\mathcal{L}_{aug}$. We report the training time of the Ego-centric Generator and the vanilla version without $\mathcal{L}_{aug}$. The quantitative results illustrate that our proposed $\mathcal{L}_{aug}$ enhances the training of the diffusion process, thus improving the semantic alignment and perceptual quality.}
  \label{tab:ablation-study-consistency-augmented-loss}
  \begin{tabular}{lccccccccc}
    \toprule
     Settings & Epoch & Training Time & Multi-View & Panorama & CLIP-Score $\uparrow$ & Sharp $\uparrow$ & Colorful $\uparrow$ & Resolution $\uparrow$ & Q-Align $\uparrow$ \\
    \midrule
    w/o $\mathcal{L}_{aug}$ & 8 & $\sim 40h$  & \checkmark & & 25.869 & 0.218 & 0.212 & 0.228 & 0.911 \\
    ours & 8 & $\sim 37h$ & \checkmark & & \textbf{25.949} & \textbf{0.219} & \textbf{0.213} & \textbf{0.228} & \textbf{0.914} \\
    \midrule
    \midrule
    w/o $\mathcal{L}_{aug}$ & 8 & $\sim 40h$ & & \checkmark & 25.686 & \textbf{0.218} & \textbf{0.216} & \textbf{0.230} & 0.809 \\
    ours & 8 & $\sim 37h$ & & \checkmark & \textbf{26.251} & 0.217 & 0.215 & 0.229 & \textbf{0.812} \\
    \bottomrule
  \end{tabular}
\end{table*}
 We further provide quantitative results for the ablation study of Ego-centric Decorator in Tab.~\ref{tab:ablation-study-consistency-augmented-loss}. 
For each configuration, we compute metrics over both the multi‑view and holistic scopes to assess semantic alignment and perceptual quality in cross-view and global settings. 
The results illustrate that the introduction of $\mathcal{L}_{aug}$ can enhance the ability to produce semantically aligned and perceptually coherent outputs in both cross‑view and global settings.

Furthermore, it is noteworthy that when $\mathcal{L}_{aug}$ is removed, output quality at the inter-view level exceeds that at the global level.
However, upon introducing consistency augmentation, the semantic and visual coherence at the global level surpasses that at the inter-view level.
This finding further demonstrates that our design is instrumental in alleviating the gradient conflicts inherent to multi-view generation.

\vspace{0.3em}
\textbf{Layout Decorator.} We compare Layout Decorator with conventional SfM methods, COLMAP~\cite{schonberger2016structure}, by directly substituting the block in \textcolor{softred}{CGGS} while keeping other modules same. 
\textcolor{softred}{To ensure a fair comparison, we evaluate COLMAP both with and without the identical camera trajectories used in our method.}
Results in Tab.~\ref{tab:ablation_sfm} demonstrate that our methods provide more reliable 3D structure for subsequent 3D Gaussian optimization.
Under the relatively sparse-view settings with little overlap across views, conventional SfM methods tend to diverge during optimization or result in suboptimal spatial structure.

\begin{table}
\centering
\caption{Ablation study of different SfM Methods. Our configuration provides more robust 3D layouts for subsequent 3D Gaussian optimization.}
\label{tab:ablation_sfm}
\resizebox{\columnwidth}{!}{
    \begin{tabular}{@{}lcccc@{}}
    \toprule
    Method & Learning Type & PSNR$\uparrow$ & SSIM$\uparrow$ & LPIPS$\downarrow$ \\
    \midrule
    COLMAP & progressive optimization & 30.133 & \textcolor{softred}{0.929} & \textcolor{softred}{0.0860} \\
    \textcolor{softred}{COLMAP (w/ pose)} & \textcolor{softred}{progressive optimization} & \textcolor{softred}{30.362} & \textcolor{softred}{0.928} & \textcolor{softred}{0.0856} \\
    \textbf{CGGS} & feed-forward Network & \textbf{37.345} & \textbf{0.997} & \textbf{0.0193} \\
    \bottomrule
    \end{tabular}
}
\end{table}

\begin{figure*}[htbp]
    \centering
    \includegraphics[width=0.95\linewidth, trim=2.0cm 1.2cm 2.0cm 2cm, clip]{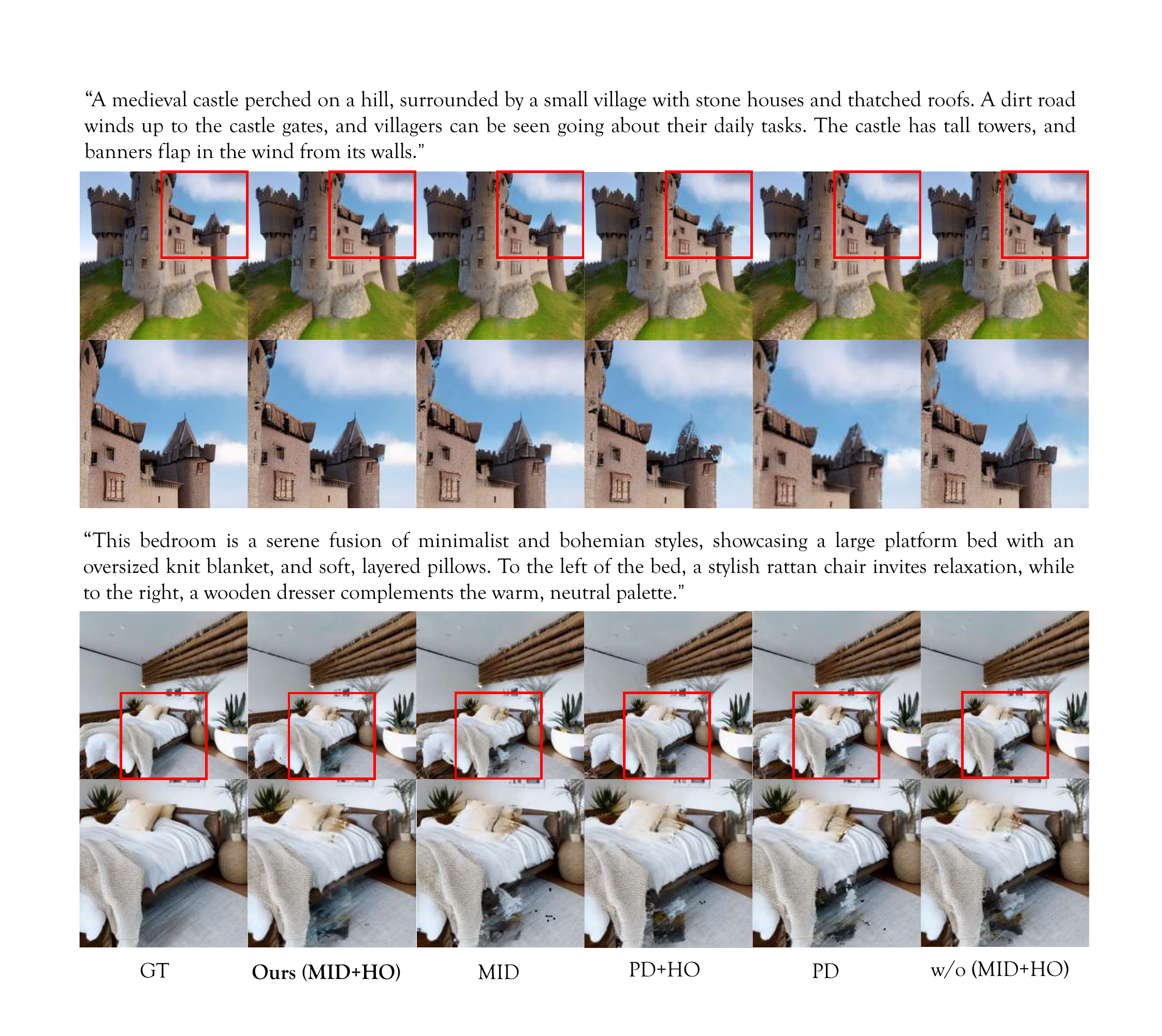}
    \caption{Ablation studies of Geometric-Refiner on MID loss and hierarchical optimization. Here we demonstrate the qualitative comparison between the ground truth with different settings, including MID+HO, MID, PD+HO, PD, and w/o (MID+HO). The comparison covers both indoor and outdoor scenes. Our design of Geometric-Refiner provides the most accurate texture recovery, with fewer blurred blocks than other settings.}
    \label{fig:ablation_study_MID}
\end{figure*}

\vspace{0.3em}
\textbf{Geometric Refiner.} The results of ablation studies on Geometric Refiner are reported in Tab.~\ref{tab:ablation-study-MID}.
Here, HO, PD, and MID briefly represent the hierarchical optimization, Pearson Depth loss, and MID loss, respectively. 
Configuration (e) is the full setting.
The quantitative comparison indicates that relying solely on the depth supervision slightly degrades the rendering visual quality, due to the stricter structure constraints.
In contrast, the combined application of hierarchical optimization and the MID loss yields substantial improvements in the geometric coherence of the 3D Gaussian primitives as well as in overall rendering performance.
Qualitative results of ablation studies on Geometric Refiner are demonstrated in Fig.~\ref{fig:ablation_study_MID}. 
The qualitative comparison demonstrates that incorporating MID and HO produces higher‑fidelity visual content that most closely approximates the ground truth.
\textcolor{softred}{
It is worth noting that this hierarchical pipeline exhibits inherent robustness to upstream inconsistencies. 
The physical correspondences in the Layout Decorator and the statistical alignment driven by the MID loss in the Geometric Refiner effectively act as filtering mechanisms, capable of repairing minor semantic or textural discrepancies present in the rendered ego-centric 2D priors.}

\begin{table}[htbp]
  \centering
  \caption{Ablation studies of Geometric Refiner on MID loss and hierarchical optimization. We explore the effectiveness of our proposed MID loss and hierarchical optimization, and compare our depth loss with the conventional Pearson Depth loss. The results below support our design choices.}
  \label{tab:ablation-study-MID}
  \begin{tabular}{ccccccc}
    \toprule
     & HO & PD & MID & PSNR$\uparrow$ & SSIM$\uparrow$ & LPIPS$\downarrow$ \\
    \midrule
    (a)      &   &   &   & 36.087 & 0.972 & 0.0231 \\
    (b)      &   & \checkmark &   & 35.883 & 0.971 & 0.0235 \\
    (c)      &   &   & \checkmark & 35.982 & 0.972 & 0.0225 \\
    (d)   & \checkmark & \checkmark &  & 36.251 & 0.971 & 0.0238 \\
    \textbf{(e)} & \checkmark &  & \checkmark & \textbf{37.345} & \textbf{0.997} & \textbf{0.0193} \\
    \bottomrule
  \end{tabular}
\end{table}

\subsection{Generalization Analysis}
\textbf{Additional Qualitative Results.} Although we simply fine-tune the CAA blocks in Ego-centric Generator with an exclusively indoor scene dataset, Matterport3D, the preservation of the other parts from stable diffusion enables our model to keep the capability of cross-domain generation.
\begin{figure*}[htbp]
    \centering
    \includegraphics[width=1\linewidth, trim=3.0cm 4.2cm 3.2cm 0cm, clip ]{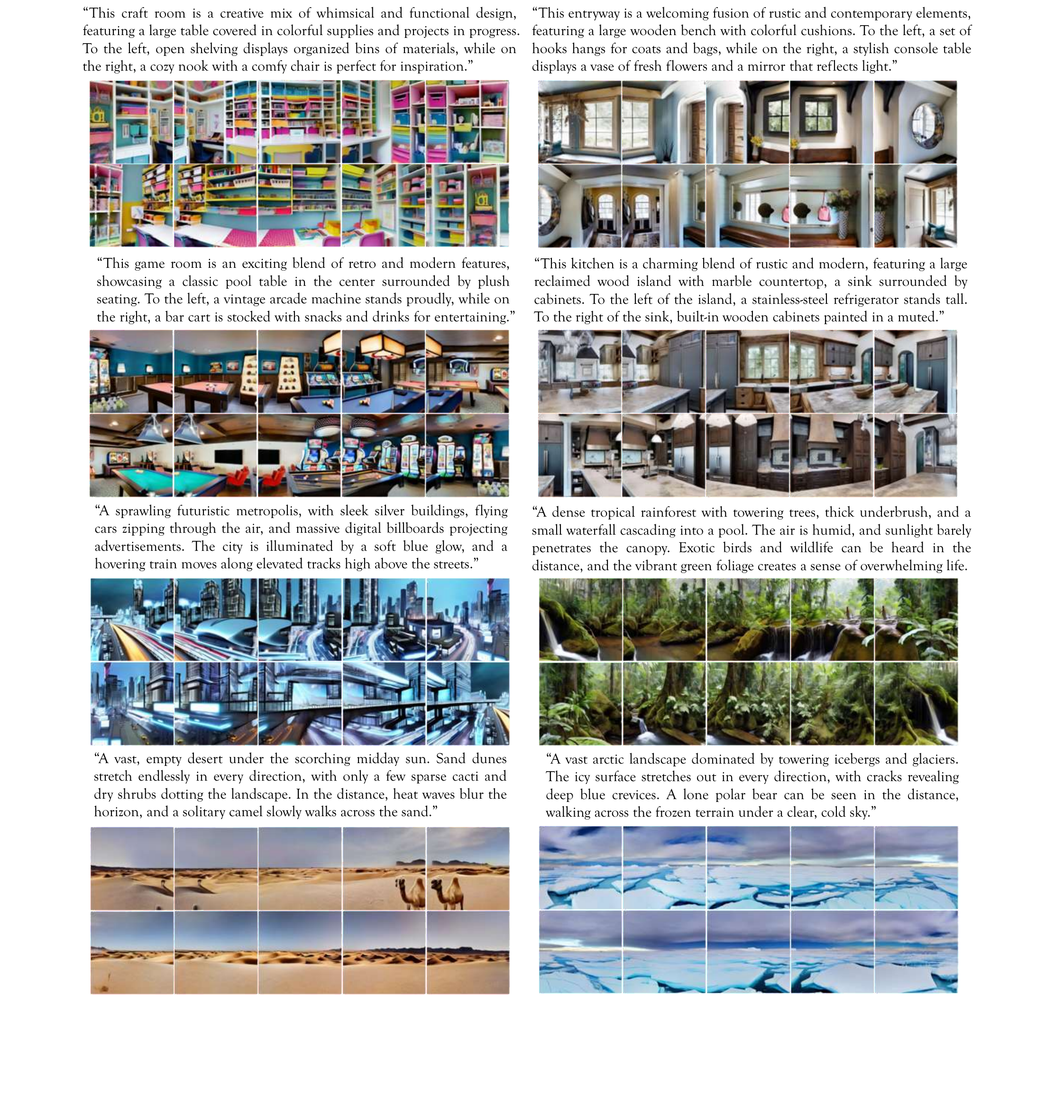}
    \caption{Additional generation results from \textcolor{softred}{CGGS}. Our work can generate richly detailed, high‑fidelity scenes with considerable diversity while ensuring cohesive semantic content and a harmonized visual style that faithfully reflects even the most intricate textual descriptions.  }
    \label{fig:supplementary_more_generation_results}
\end{figure*}
And the data from RealEstate-10k and CO3Dv2 provide powerful prior knowledge of real-world structure, therefore promoting the derivation of initial layout from Layout Decorator, further improving the 3D structure representation and visual details in the Geometric Refiner process. 
\begin{figure}[htbp]
    \centering
    \includegraphics[width=1\linewidth, trim=3.6cm 2cm 18.5cm 2.0cm, clip ]{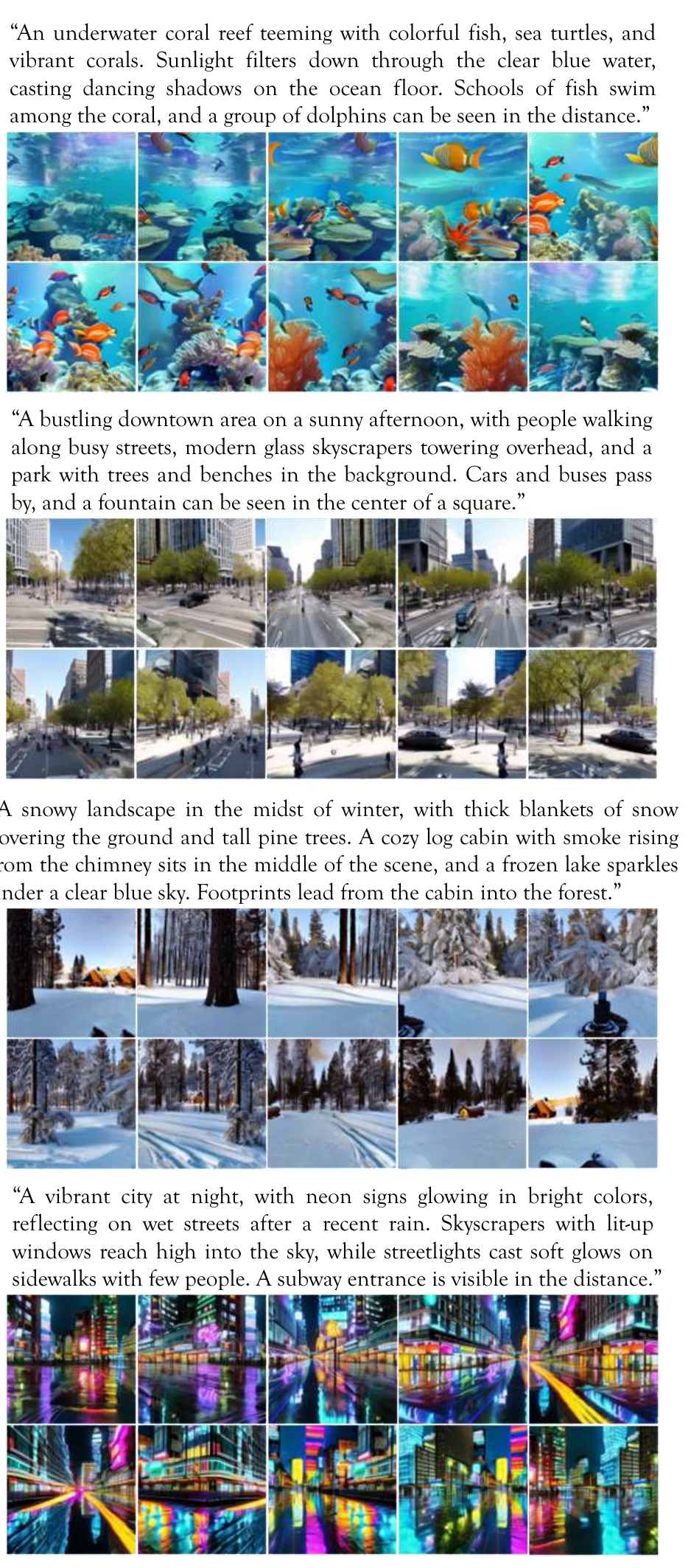}
    \caption{Additional generation results from \textcolor{softred}{CGGS}. Our work can generate richly detailed, high‑fidelity scenes with considerable diversity while ensuring cohesive semantic content and a harmonized visual style that faithfully reflects even the most intricate textual descriptions.  }
    \label{fig:supplementary_more_generation_results_2}
\end{figure}

We provide more generation results in Fig.~\ref{fig:supplementary_more_generation_results} and Fig.~\ref{fig:supplementary_more_generation_results_2}.
The content covers various indoor and outdoor scenarios. These visual results demonstrate the impressive ability of \textcolor{softred}{CGGS} on novel view synthesis from complex textual descriptions and provide compelling evidence for the effectiveness of our full pipeline. 
Specifically, they demonstrate that the Ego-centric Generator produces semantically faithful 2D priors, while the Layout Decorator successfully reconstructs coarse 3D layouts even in challenging ego-centric conditions. 
The Geometric Refiner further enhances geometric fidelity by enforcing structural consistency through hierarchical optimization, ultimately resulting in high-quality, cross-view-consistent reconstructions.

\begin{table}
\centering
\caption{Quantitative comparison on out-of-domain scenes. Our CGGS achieves competitive image quality and strong semantic alignment, and it surpasses previous methods on overall generative performance, indicated by the highest Q-Align score.}
\label{tab:quantitative comparisons on OOD}
\resizebox{\columnwidth}{!}{
    \begin{tabular}{@{}lccccc@{}}
    \toprule
    \multirow{2}[2]{*}{Method} & \multirow{2}[2]{*}{CLIP Score$\uparrow$} & \multicolumn{3}{c}{CLIP-IQA} & \multirow{2}[2]{*}{Q-Align$\uparrow$} \\
    \cmidrule(lr){3-5}
     & & Sharp$\uparrow$ & Color$\uparrow$ & Resolution$\uparrow$ &  \\
    \midrule
    Text2Room & 24.14 & 0.210 & 0.220 & 0.224 & 0.669 \\
    LucidDreamer & \textbf{25.84} & 0.207 & \underline{0.222} & 0.221 & 0.676 \\
    Director3D & 22.95 & \underline{0.218} & \textbf{0.225} & 0.223 & 0.475 \\
    DreamScene360 & 25.36 & 0.217 & 0.216 & \textbf{0.232} & \underline{0.791} \\
    \textbf{CGGS} & \underline{25.80} & \textbf{0.220} & 0.214 & \underline{0.227} & \textbf{0.820} \\
    \bottomrule
    \end{tabular}
}
\end{table}
\textbf{Domain-Free Generation \& Generalization.}
An important observation is that the diverse examples presented in our paper include many out-of-domain scenarios not seen during training, such as urban scenes with pedestrians and vehicles, as well as scenes dominated by humans, animals, and other biological entities.
\textcolor{softred}{We further conduct quantitative evaluation on generation performance between these methods, with 4 out-of-domain scenes---camels in the desert, people in the market, creatures underwater, pedestrians in urban daytime.}
\textcolor{softred}{As illustrated in Tab.~\ref{tab:quantitative comparisons on OOD}, CGGS competes with existing methods in image quality, and surpasses them in semantic alignment and the overall generation quality.}
This demonstrates that \textcolor{softred}{CGGS} exhibits notable generalization ability, generating semantically faithful 3D scenes from textual descriptions with accurate details and consistent spatial structures. 
These results indicate that our proposed \textcolor{softred}{CGGS} effectively tackles the challenges of ego-centric 3D generation and reconstruction, while preserving strong diversity and generalization performance.

\textbf{Limitations and Future Work.} 
Despite strong performance in text‑driven ego-centric 3D synthesis, \textcolor{softred}{CGGS} remains limited by per‑scene optimization, which increases computation time and hampers generalization. 
Future work will explore dynamic scene synthesis under ego-centric settings and visual language navigation within the generated environments via LLMs.

\textbf{Broader Impacts.} This paper introduces a framework aiming at improving semantic alignment and overall perceptual coherence in ego-centric 3D scene generation.
Due to the inherent viewpoint characteristics of the generated scenes, the system holds considerable promise for synthetic data generation in the autonomous driving domain. 
Moreover, the proposed reconstruction enhancements may substantially improve the fidelity of imagery reconstructed from vehicle‑mounted cameras.
Such synthetic reconstructions may exacerbate privacy infringements or be weaponized to create convincing digital forgeries, posing substantial risks to societal security.
\section{Conclusion}
\label{sec:conclusion}
In this work, we propose \textbf{\textcolor{softred}{CGGS},} a novel text-to-3D framework designed to address the challenges of ego-centric 3D scene generation. 
Ego-centric Generator is introduced to synthesize high‑fidelity 2D content aligned with textual prompts. 
Layout Decorator is proposed to produce reliable initialization from ego-centric 2D priors, then Geometric Refiner is leveraged for further optimization, using 3D Gussians as scene representation. 
We validate the effectiveness of our methods via comprehensive experiments.
We believe that \textcolor{softred}{CGGS} not only enhances current ego-centric 3D scene generation approaches but also paves the way for more diverse and realistic 3D content creation and virtual environment exploration.

\newpage

{
\small
\bibliographystyle{IEEEtran}
\bibliography{main}
}

\end{document}